\documentclass{article}

\usepackage{arxiv}
\usepackage[utf8]{inputenc} 
\usepackage[T1]{fontenc}    
\usepackage{hyperref}       
\usepackage{url}            
\usepackage{booktabs}       
\usepackage{amsfonts}       
\usepackage{amsmath}        
\usepackage{nicefrac}       
\usepackage{microtype}      
\usepackage{lipsum}
\usepackage{graphicx}
\usepackage{subcaption}
\graphicspath{ {./images/} }

\usepackage{xcolor}

\title{Semantic Saliency from Multi-Modal Large Language Model Scene Understanding Maps}

\author{
 Ziqi Wen \\
  Department of Computer Science\\
  University of California, Santa Barbara\\
  Santa Barbara, CA 93117 \\
  \texttt{ziqiwen@ucsb.edu}\\
  \And
Shravan Murlidaran \\
  Department of Psychological and Brain Sciences\\
  University of California, Santa Barbara\\
  Santa Barbara, CA 93117 \\
   \And
 Jonathan Skaza \\
  Graduate Program in Dynamical Neuroscience\\
University of California, Santa Barbara\\
  Santa Barbara, CA 93117
  \And
 William Wang \\
  Department of Computer Science\\
  Department of Electrical and Computer Engineering\\
University of California, Santa Barbara\\
  Santa Barbara, CA 93117
  \And
 Miguel P Eckstein \\
  Department of Psychological and Brain Sciences\\
  Department of Computer Science\\
  Department of Electrical and Computer Engineering\\
  Graduate Program in Dynamical Neuroscience\\
University of California, Santa Barbara\\
  Santa Barbara, CA 93117
}

\begin{document}

\maketitle

\begin{abstract}
Image-computable low-level image saliency has shaped perceptual psychology and computer science, but is limited by its inability to capture human high-level cognitive factors that influence attention and eye movements.  Here, we propose a new fully automated method to compute semantic saliency by quantifying each object's contribution to scene comprehension without fitting any parameters to human data: the Automated Scene Understanding Map (AUTO-SUM). AUTO-SUM automatically segments and removes individual objects from a scene, uses multimodal large language models (MLLMs) to describe the original and modified scenes, and measures the semantic similarity between these descriptions using text embeddings. We show that AUTO-SUM can predict which object is most important in a human-measured scene understanding map (H-SUM) estimated using human scene descriptions. We also show that AUTO-SUM can predict the object most fixated by human observers during free viewing and scene description tasks better than a low-level saliency model (GBVS, Itti-Koch), a saliency model trained based on human meaning ratings (Meaning Map), and comparable to DeepGaze, which is trained on human fixation data. We contend that AUTO-SUM can be used as a semantic saliency model that complements lower-level saliency models. Code are available at \href{https://github.com/starsky77/AUTO-SUM-Source-Code}{https://github.com/starsky77/AUTO-SUM-Source-Code}
\end{abstract}

\section{INTRODUCTION}

Visual saliency has been an influential concept for the past 25 years.  It aims to capture properties of image regions or objects that make them visually pop out and attract attention. Image-computable visual saliency was originally proposed as a predictor of human eye movements during free viewing of eye movements and visual search \cite{itti2000saliency, parkhurst_modeling_2002, bruce2005saliency, bruce_saliency_2009, bruce_computational_2015, borji2018saliency}. The original visual saliency model computes local luminance contrasts, orientations, and colors present in different regions of a scene using center-surround operations \cite{itti1998model, harel2006graph}. Subsequent models extended the saliency definition to include global context, sparse encoding, and statistical surprise \cite{zhang2008sun, goferman2011context}. However, studies have shown the importance of higher-level cognition in influencing attention and eye movement patterns for tasks such as search \cite{eckstein2017probabilistic, henderson_visual_2007, castelhano2007initial, zelinsky_theory_2008, einha2008task, malcolm2009effects, malcolm2010combining, torralba_contextual_2006, koehler2014saliency, hayhoe2017vision,koehler_beyond_2017,vo2019reading, han2023head, eckstein2015optimal}, visual navigation \cite{hayhoe2005eye}, face identification \cite{peterson2012looking}, and motor actions \cite{de2021functional, fooken2021role, hayhoe2022visual}. These task-related cognitive goals are typically absent in low-level saliency models. 

\hfill

Even when given no specific task or instructions (free viewing), humans fixate on objects and meaningful regions more than on low-level salient regions. To capture the high-level component of attention and eye movements during free viewing, investigators have adopted data-driven machine learning models trained on human fixation (\cite{judd2009learning, kruthiventi2015deepfix, pan2017salgan, linardos2021deepgaze}), developed models that compute semantic object relationships (\cite{hayes2021looking}) or use psychophysics to empirically measure observer judgments of the meaningfulness of local image regions cropped from scenes (\cite{henderson_meaning_2018}). Very recently, Murlidaran and Eckstein \cite{murlidaran2025eye} have demonstrated that a scene understanding map (SUM), which quantifies the contribution of an object to the understanding of a scene (as measured by scene descriptions), predicts fixation regions better than DeepGaze or Saliency during free viewing and scene understanding. The SUMs are estimated by a group of humans rating the similarity between descriptions provided by humans to a scene and its digitally manipulated versions, where each version has one object digitally removed. The lower the similarity, the higher the contribution of that object to scene understanding. The scalar ratings measuring the impact of the removal of an object on scene understanding can be mapped back to the image to create the human scene understanding map, H-SUM (see Figure \ref{fig:panel1}a). Although the SUM procedure is a useful method for measuring regions that attract the most attention (semantic saliency), it is currently labor-intensive and time-consuming, preventing the scaling up of the approach to large image databases. 

In this context, the goal of the present study is to introduce a new methodology, AUTO-SUM, a scene understanding map estimator that automates the procedure of the SUMs.   Advances in image detection and segmentation \cite{maaz2023video, kirillov2023segment}, along with advances in multimodal large language models (MLLMs) that can reason and respond like humans \cite{jiang2023active, achiam2023gpt, reid2024gemini} have provided a unique opportunity to automate various components of the SUM generation process. AUTO-SUM automates the following components: 1) The Object Removal Stage (OR), 2) The Scene Descriptions Stage (SD), and 3) The Similarity Rating Stage (SR). We evaluate the agreement between the AUTO-SUM results and the human-created SUM (H-SUM), and show that AUTO-SUM can better predict the most fixated object compared with the low-level saliency map (GBVS\cite{harel2006graph}, Itti-Koch\cite{itti1998model}), saliency model trained to predict the human meaning map \cite{henderson2017meaning,hayes2025deepmeaning}
and is comparable to the deep learning method trained on human fixation data (DeepGaze IIE\cite{linardos2021deepgaze}).


\section{RESULTS}
 
\begin{figure}[ht!]
  \centering
  \includegraphics[width=0.82\linewidth]{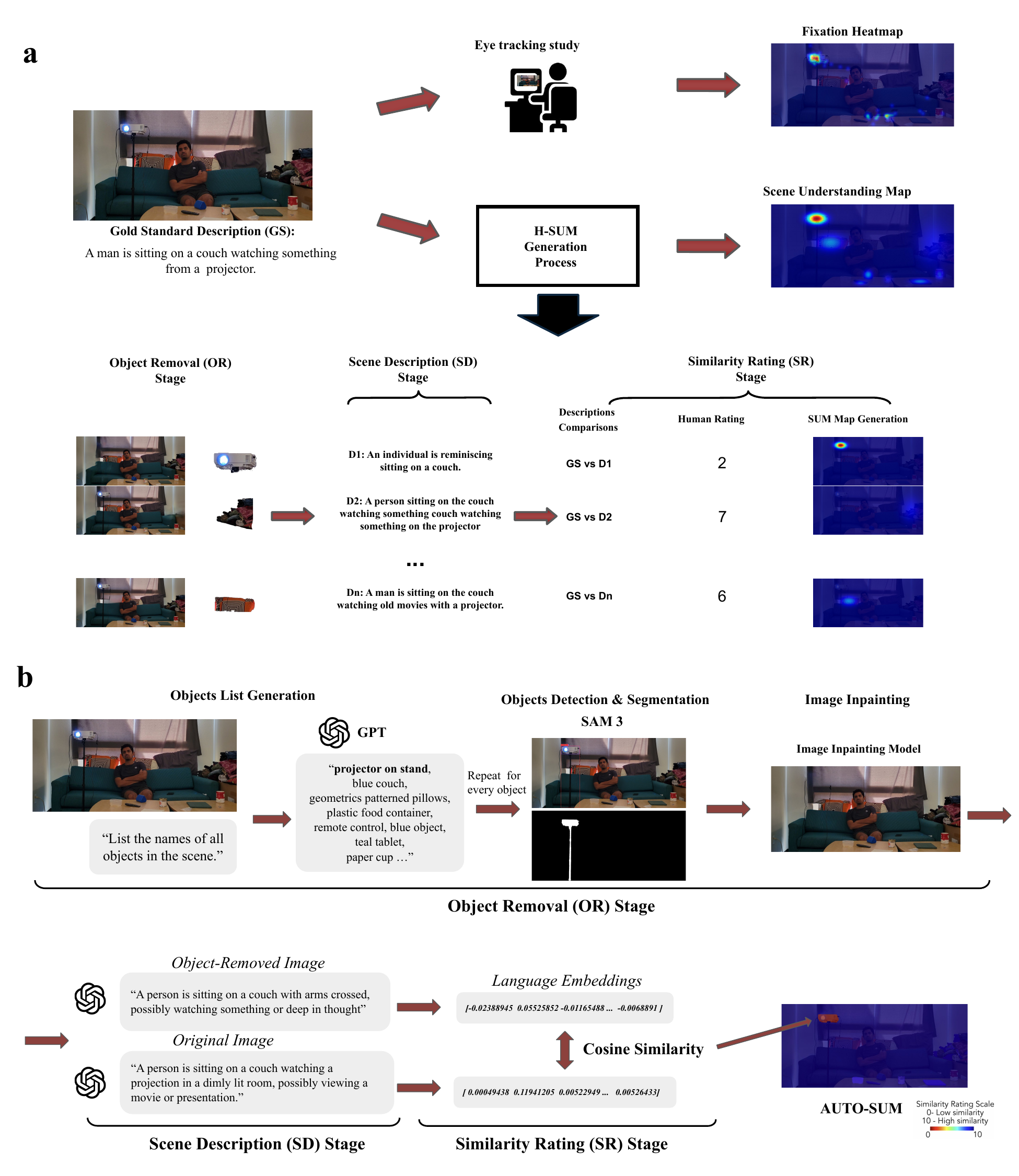}
  \caption{\textbf{Generation of human and automated scene understanding maps.}
  (a) The original Human scene understanding map (H-SUM) procedure from Murlidaran and Eckstein~\cite{murlidaran2025eye}. In the Object Removal (OR) stage, objects are digitally removed from the scene one at a time. In the Scene Description (SD) stage, human participants describe the original scene (gold standard description, GS) and each object-removed version (D1, \dots, Dn). In the Similarity Rating (SR) stage, human raters judge the similarity between the GS and each description of a manipulated scene; the lower the similarity, the larger the contribution of the removed object to scene understanding, and the object's region is weighted accordingly to produce the scene understanding map. The fixation heatmap from an eye-tracking study of the same scene is shown for comparison. In the present study, the H-SUM uses human descriptions (SD stage) but the same automated OR and SR stages as AUTO-SUM.
  (b) AUTO-SUM. OR stage: an MLLM is prompted to list all objects in the scene; each object name is passed to a text-prompted detection and segmentation model to obtain the object's mask, and an image inpainting model removes the masked object. This is repeated for every listed object. SD stage: an MLLM describes the original image and each object-removed image. SR stage: the descriptions are encoded by a language-embedding model, and the cosine similarity between the embeddings of the original and the manipulated scene descriptions replaces the human similarity rating. Each object's mask is weighted by its similarity-based score to produce the AUTO-SUM.}
  \label{fig:panel1}
\end{figure}

Figure~\ref{fig:panel1} summarizes how scene understanding maps are produced with human participants (H-SUM; Fig.~\ref{fig:panel1}a) and automatically (AUTO-SUM; Fig.~\ref{fig:panel1}b). Both procedures quantify the contribution of an object to scene understanding in the same way: the object is removed from the scene, the manipulated scene is described, and this description is compared with the description of the original scene. The less similar the two descriptions, the more the removed object contributed to scene understanding, and the higher the value assigned to the object's region in the map. In the original H-SUM procedure \cite{murlidaran2025eye}, each of the three stages is carried out by humans (Fig.~\ref{fig:panel1}a). AUTO-SUM replaces them with automated components (see Methods): in the Object Removal (OR) stage, an MLLM lists the objects in the scene, a segmentation model produces a mask for each listed object, and an inpainting model removes the masked object; in the Scene Description (SD) stage, an MLLM describes the original image and each object-removed image; in the Similarity Rating (SR) stage, the descriptions are encoded by a language-embedding model, and the cosine similarity between the embeddings of the original and the manipulated scene descriptions serves as a proxy for the human similarity rating. No component of AUTO-SUM is trained or fitted on human SUMs or on eye-movement data. We report two variants of the pipeline, AUTO-SUM-C (closed-source) and AUTO-SUM-O (open-source), which differ only in the inpainting model used in the OR stage and the MLLM used in the SD stage. AUTO-SUM-C uses Google Nano Banana 2 for inpainting and GPT-5.4 for scene description; AUTO-SUM-O uses the Mask-aware Transformer (MAT)~\cite{li2022mat} for inpainting and Ovis2~\cite{lu2024ovis} for scene description.

To isolate the role of scene descriptions, the H-SUM used as ground truth in the present study keeps only the SD stage human. The OR stage reuses the object-removed images generated by AUTO-SUM-C, and the SR stage uses the same embedding-based similarity as AUTO-SUM, which agrees with human similarity ratings about as well as human raters agree with each other~\cite{murlidaran2025eye} (see Methods). H-SUM and AUTO-SUM therefore differ only in whether the scene descriptions come from humans or from an MLLM.
 
We compared AUTO-SUM with four image-computable models of fixation (see Methods): two bottom-up saliency models based on low-level image features (Itti-Koch\cite{itti1998model} and GBVS\cite{harel2006graph}); DeepMeaning\cite{hayes2025deepmeaning}, which predicts human ratings of the semantic meaning of local image patches without access to eye-movement data; and DeepGaze IIE\cite{linardos2021deepgaze}, which is trained on human fixations and serves as a state-of-the-art reference. Because these models produce pixel-level maps, whereas SUMs assign a value to each object, we converted each model's map into object-level scores by taking either the maximum or the mean of the map within each object's segmentation mask, denoted (max) and (mean), respectively. The segmentation masks used here are the same as in the AUTO-SUM procedure.

\subsection*{Automated Scene Understanding Map predicts the Human Scene Understanding Map}
 
\begin{figure}[ht!]
  \centering
  \includegraphics[width=\linewidth]{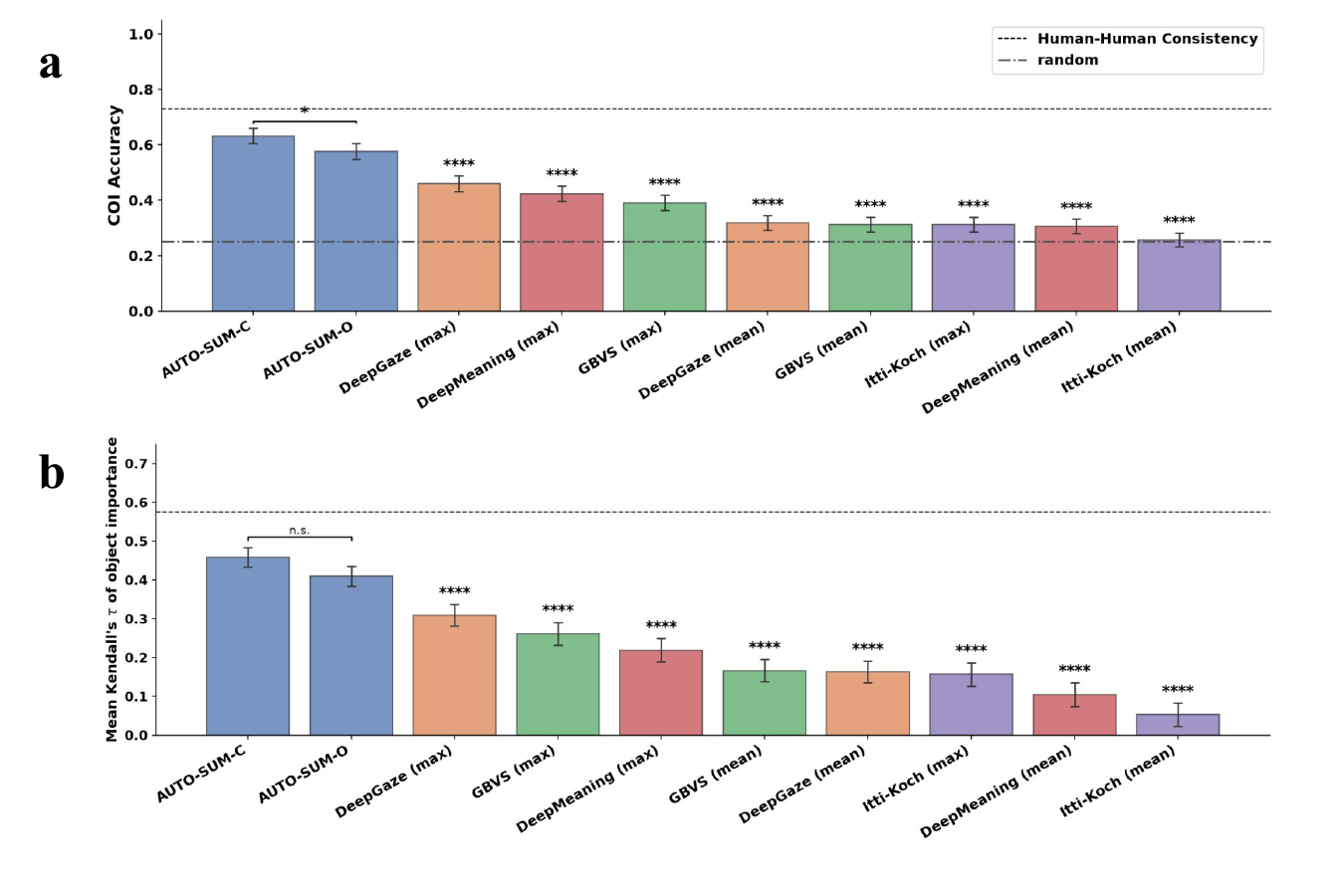}
  \caption{\textbf{AUTO-SUM predicts the Human Scene Understanding Map (H-SUM) on 307 images from the OSIE dataset.}
  (a) Critical Object Identification (COI) accuracy: the proportion of images in which a model's highest-scoring object matches the object with the largest impact on human scene understanding. The ground truth is the H-SUM of one group of five participants (computed from human scene descriptions; see Methods). Human--human consistency (dashed line) is the accuracy with which the H-SUM of the other five participants predicts the ground truth. The dash-dotted line shows the expected accuracy of selecting one object at random from each scene.
  (b) Rank-order correlation between the H-SUM and the SUM predicted by each model, measured as Kendall's $\tau$ between the two object-importance rankings and averaged across images. The dashed line shows human--human consistency.
  (max) and (mean) denote object scores computed as the maximum or the mean of a model's map within the object mask. All the error bars are obtained using the std of bootstrap samples of images. $* = p<.05$; $** = p<.01$; $*** = p<.001$; $**** = p<.0001$ from one-tailed bootstrap test.}
  \label{fig:panel2}
\end{figure}
 
We first asked whether AUTO-SUM recovers the object-level structure of human scene understanding maps. We evaluated all models against H-SUMs on 307 natural images from the Object and Semantic Images and Eye-tracking (OSIE) dataset \cite{xu2014predicting} using two complementary metrics. Critical Object Identification (COI) accuracy is the proportion of images for which a model's highest-scoring object matches the object whose removal most altered human scene descriptions, that is, the object with the highest score from the H-SUM. Because COI accuracy considers only the single most important object, we also assessed how well each model orders all objects in a scene by computing Kendall's rank correlation coefficient ($\tau$) between the object-importance ranking given by the H-SUM and the ranking given by the model, averaged across images. To estimate the ceiling that can be expected given human variability, we computed human--human consistency by using the H-SUM of one group of five participants as the ground truth and the H-SUM of a separate group of five participants as the predictor. For COI accuracy, the floor is the expected accuracy of selecting one object at random from each scene.
 
\begin{figure*}[t]
    \centering
    \begin{subfigure}[t]{0.49\textwidth}
        \centering
        \includegraphics[width=\linewidth]{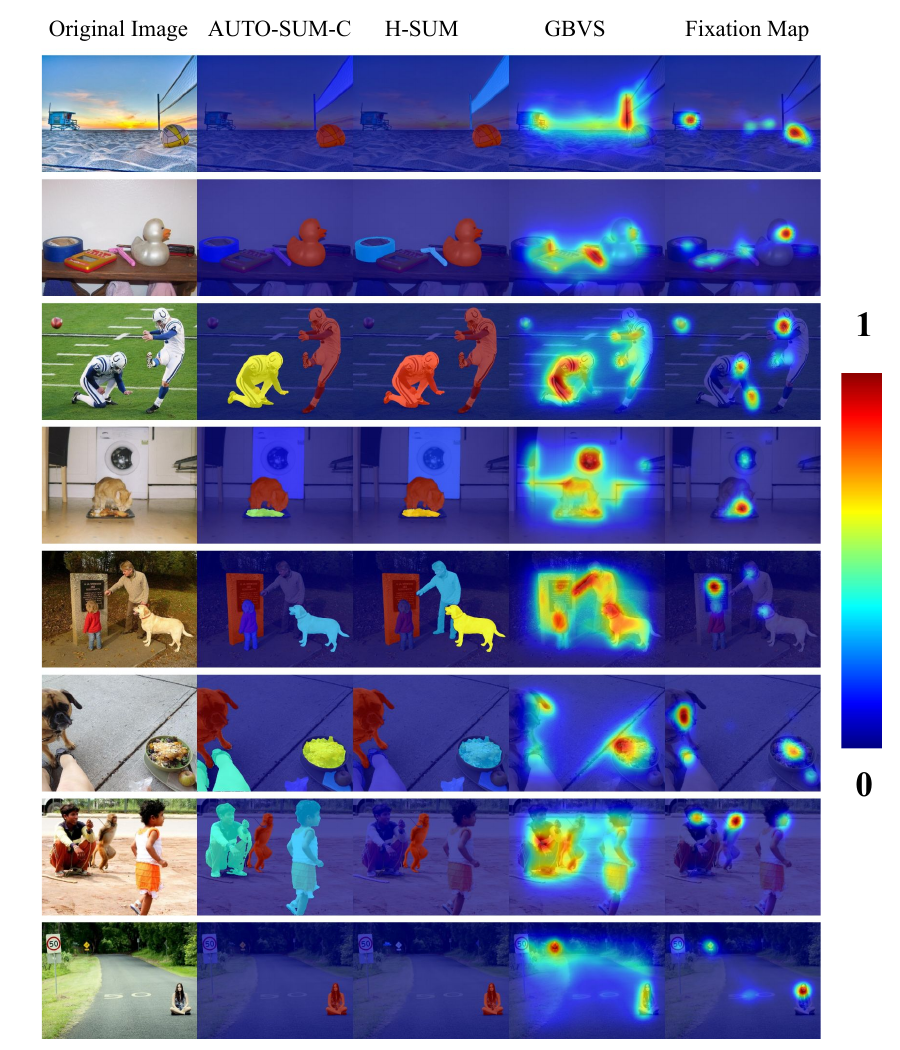}
        \caption{AUTO-SUM-C and H-SUM agree.}
        \label{fig:examples_same}
    \end{subfigure}
    \hfill
    \begin{subfigure}[t]{0.49\textwidth}
        \centering
        \includegraphics[width=\linewidth]{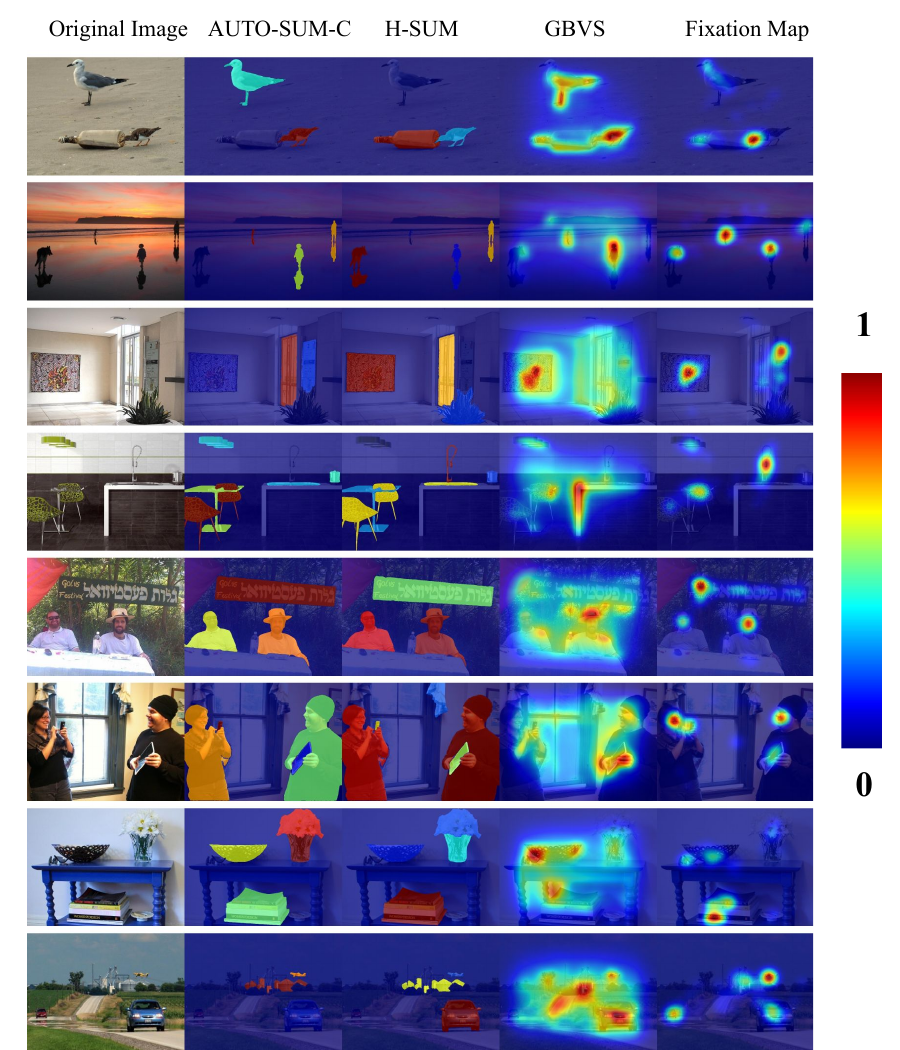}
        \caption{AUTO-SUM-C and H-SUM disagree.}
        \label{fig:examples_differ}
    \end{subfigure}
    \caption{\textbf{Qualitative comparison of scene understanding maps, low-level saliency (GBVS), and human fixations.} Each row shows the original image, AUTO-SUM-C, H-SUM, the GBVS saliency map, and the human fixation map; all maps are normalized to $[0, 1]$. (a) Scenes in which AUTO-SUM-C and H-SUM identify the same most critical object. (b) Scenes in which the two maps assign the highest importance to different objects. In both cases, GBVS often peaks on high-contrast regions that humans rarely fixate, such as the volleyball net pole and the washing machine in (a) or the countertop edge in (b). In contrast, AUTO-SUM-C and H-SUM more often assign high importance to the objects that attract the most fixations, such as the volleyball, the cat and its food, and the small bird next to the bottle.}
    \label{fig:qualitative_examples}
\end{figure*}

Both AUTO-SUM variants identified the critical object well above chance (Fig.~\ref{fig:panel2}a; AUTO-SUM-C:$Acc=0.63$, AUTO-SUM-O $Acc=0.58$), approaching human--human consistency ($Acc=0.73$) the most compared with all other metrics, though there is still a gap between AUTO-SUM and H-SUM. Fig.\ref{fig:examples_same} shows the examples when AUTO-SUM-C consists with H-SUM, and Fig.\ref{fig:examples_differ} contains the cases when they disagree. All comparison models were significantly less accurate than AUTO-SUM-C ($p<.0001$). The best comparison model was DeepGaze (max) ($Acc=0.46$), followed by DeepMeaning (max) ($Acc=0.42$) and GBVS (max) ($Acc=0.39$), whereas the remaining models performed close to chance.
 
The same pattern held for the ranking of all objects in a scene (Fig.~\ref{fig:panel2}b). AUTO-SUM-C and AUTO-SUM-O, which did not differ significantly from each other, came closest to human--human consistency. All comparison models showed significantly lower rank correlations with the H-SUM than AUTO-SUM-C ($p<.0001$), ranging from DeepGaze (max) to Itti-Koch (mean). Across both metrics, max aggregation outperformed mean aggregation for every comparison model.

\subsection*{Automated Scene Understanding Map predicts the most fixated objects}
 
\begin{figure}[ht!]
  \centering
  \includegraphics[width=\linewidth]{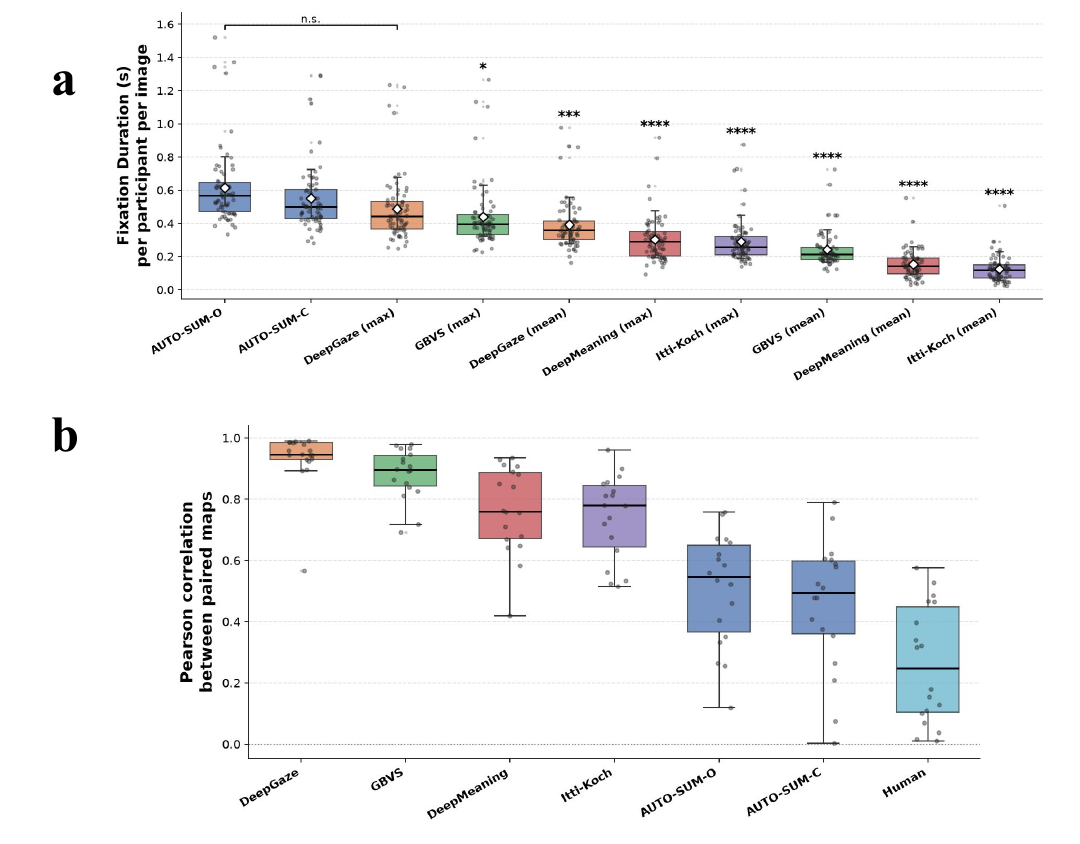}
  \caption{\textbf{Evaluation of each model's ability to predict human fixations on 36 Winoground images}. $* = p<.05$; $** = p<.01$; $*** = p<.001$; $**** = p<.0001$ from one-tailed bootstrap test. (a) Mean Fixation Duration (s) of the highest-scoring object under each model. The box is calculated based on the quartiles of the subjects' data (n=45), the middle line indicates the median. The diamond is the per-participant mean and the error bars are obtained using the 2.5\%–97.5\% percentile CI of bootstrap samples of both subjects and images. Each data point here represents a subject (b) Within-pair similarity of maps for the 18 Winoground image pairs. For each pair, we computed the Pearson correlation between the maps of its two images, separately for each model and for human fixation maps. Each data point represents one Winoground pair. Higher values indicate that a map changes less between the two images of a pair. Boxes show the interquartile range, the middle line indicates the median, and whiskers extend to the most extreme data points within 1.5$\times$ the interquartile range.}
  \label{fig:panel3}
\end{figure}
 
\begin{figure}[ht!]
  \centering
  \includegraphics[width=\linewidth]{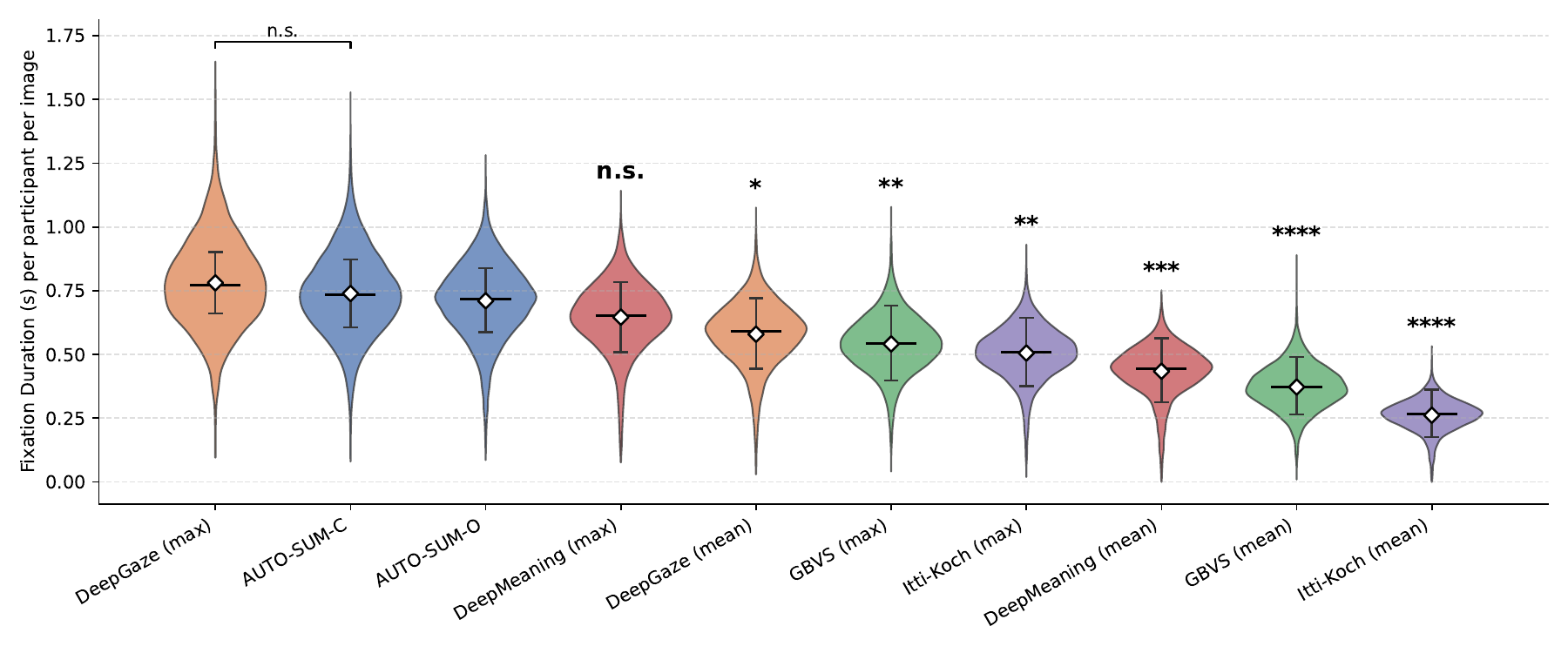}
  \caption{\textbf{AUTO-SUM predicts human fixations on 40 images from \cite{linka2025protracted}.} Fixation duration (ms) on the highest-scoring object of each model, per participant per image. Violins show the distribution across 6,720 participants. The diamond is the per-participant mean, and the error bars are the 2.5\%–97.5\% percentile CI from bootstrap samples of both subjects and images. $* = p<.05$; $** = p<.01$; $*** = p<.001$; $**** = p<.0001$ from one-tailed bootstrap test.}
  \label{fig:panel4}
\end{figure}

We next asked whether the objects that AUTO-SUM identifies as critical for scene understanding are also the objects that human observers look at. For each model, we selected the highest-scoring object in each image and measured the time each participant spent fixating that object (fixation duration per participant per image). A model that better predicts which object attracts human gaze should select objects with longer fixation durations. We performed this analysis on two eye-tracking datasets: 18 Winoground images pairs (36 images in total) viewed by 45 participants under scene description tasks from \cite{murlidaran2025eye}, and 40 images viewed by 6,720 participants under free-viewing tasks from \cite{linka2025protracted}, where the Winoground images pair contains two complex scenes which designed to be visually similar, while has semantically different scene description.
 
On the Winoground images (Fig.~\ref{fig:panel3}a), the objects ranked highest by AUTO-SUM-O received the longest fixation durations, followed by those of AUTO-SUM-C and DeepGaze (max). The difference between AUTO-SUM-O and DeepGaze (max) was not significant. The objects selected by AUTO-SUM-O were fixated significantly longer than those selected by GBVS (max), DeepGaze (mean), and all remaining models, including DeepMeaning (max) and Itti-Koch (max).
 
The results generalized to the dataset from Linka et al.\cite{linka2025protracted} (Fig.~\ref{fig:panel4}). On this dataset, DeepGaze (max) selected the objects with the longest fixation durations, but it did not differ significantly from AUTO-SUM-C , and AUTO-SUM-O performed similarly. DeepMeaning (max) did not differ significantly from AUTO-SUM-C, whereas DeepGaze (mean), GBVS (max), Itti-Koch (max) , DeepMeaning (mean), GBVS (mean) , and Itti-Koch (mean) all selected objects that were fixated significantly less than those selected by AUTO-SUM-C. As in the H-SUM analysis, max aggregation outperformed mean aggregation for every comparison model on both eye-tracking datasets. Thus, although AUTO-SUM is never exposed to eye-movement data, the objects it identifies as critical to scene understanding attract as much gaze as those selected by DeepGaze, a model trained to predict fixations, and more gaze than those selected by low-level saliency models.

As reported by \cite{murlidaran2025eye}, the two images within a Winoground pair are visually similar but has different human fixation patterns, consistent with their different scene descriptions. We therefore asked whether the model maps show a comparable sensitivity to this semantic difference. For each Winoground pair, we computed the Pearson correlation between the maps a given model produced for the two images and compared it with the corresponding correlation between the human fixation maps (Fig.~\ref{fig:panel3}b). A high within-pair correlation indicates that a map changes little between two images that are visually similar but semantically different. Maps from DeepGaze, GBVS, DeepMeaning and Itti-Koch were highly correlated within pairs (median $r=0.95, r=0.90, r=0.76, r=0.78$, respectively), suggesting that they mainly reflect the shared visual content of the two images. In contrast, both AUTO-SUM variants yielded markedly lower within-pair correlations, (median $r=0.55$ for AUTO-SUM-O and median $r=0.50$ for AUTO-SUM-C). Among all models, these values were the closest to those of human fixation maps (median $r=0.25$), although they remained higher. These results indicate that, like human fixations, AUTO-SUM maps change more between scenes whose semantic difference exceeds their visual difference.

\section{DISCUSSION}

The goal of the paper was to propose an image-computable automated model that can predict which parts of an image are most critical to the understanding of the scene.  The AUTO-SUM generates such a scene understanding map where the value assigned to each object corresponds to a prediction of its contribution to the comprehension of the scene. 
The premise is that such maps will predict which objects attract attention and eye movements when humans view the scene without a task (free viewing).  Recent work \cite{murlidaran2025eye} shows empirical evidence supporting this hypothesis. The AUTO-SUM can be used as a Semantic Saliency model, unlike low-level saliency models \cite{itti2000saliency,bruce2005saliency, borji2018saliency, zhang2013saliency}, which capture the semantic role of each object in the observers' comprehension of the scene.
Our current results show that AUTO-SUM can be used as a proxy for a human-generated SUM (H-SUM), although there is room to improve the model. 

Fig.~\ref{fig:examples_differ} shows examples in which AUTO-SUM-C disagrees with H-SUM. Because humans and models described the same object-removed images, these divergences must arise from differences between MLLM-generated and human-generated scene descriptions. Deep neural networks trained on human-generated data can nonetheless rely on different visual cues than humans \cite{geirhos2018imagenet, geirhos2020shortcut}, and vision-language models in particular often overlook visual details that humans readily notice \cite{tong2024eyes}. In addition, MLLMs can infer a removed object from the surrounding context, or even describe it as if it were still present \cite{rohrbach2018object}, which reduces the semantic shift caused by its removal. These differences are properties of the MLLMs themselves rather than of the AUTO-SUM pipeline, and they set an upper bound on how closely AUTO-SUM can approximate H-SUM. Closing this gap will require a better understanding of how MLLM scene comprehension diverges from human scene comprehension, as well as methods that align the two more closely.

Beyond the limitations of current MLLMs, the object-removal approach itself has conceptual limitations that affect H-SUM and AUTO-SUM alike. The first is redundant object errors. When several objects in a scene convey the same information about the scene's meaning, removing any one of them does not substantially change the description, because the remaining objects still support the same interpretation. Each object then receives a low score, and it appears that none of them matter for scene understanding when, in fact, they all do, only redundantly. For example, in a scene of a birthday party with a cake, candles, and party hats, removing only the cake may still leave the scene described as a birthday party. One possible solution is to remove semantically related objects together rather than one at a time. Objects that provide similar information about a scene are likely to be semantically related, so grouping objects by semantic similarity, for example with concept maps \cite{hayes2021looking}, and removing one group at a time would capture the joint contribution of redundant objects.

The second conceptual limitation is that removing an object can produce an atypical scene whose description changes for reasons unrelated to the object's role in the scene. Consider an image of two people fighting next to a table with bottles and plates, described as ``two people fighting.'' If the table is removed but the bottles and plates are not, the edited image shows objects floating in mid-air. A description of this edited scene might mention the floating objects, so that it differs substantially from the original description, and the table would be scored as critical to scene comprehension. This score would be an artifact, because the table is irrelevant to the interpretation of the scene as two people fighting. Our human validation step verified that each target object was removed, but it did not assess whether the resulting scene remained physically plausible. Future versions of the pipeline should therefore include automated checks that object removal does not produce implausible scenes, for example by removing physically dependent objects together.

An important distinction is that AUTO-SUM is not trained on human eye-movement data, unlike DeepGaze \cite{kummerer_deep_2015, kummerer_deepgaze_2019, kummerer2022deepgaze} and related approaches \cite{judd2009learning, pan2017salgan}. Even without such training, AUTO-SUM predicts the most fixated object comparably to or better than DeepGaze, and better than the low-level saliency model GBVS. In addition, AUTO-SUM is interpretable, because it explicitly computes the contribution of each object to scene comprehension. 

Finally, the current AUTO-SUM approach does not account for individual \cite{kollenda2025individual} or cross-cultural \cite{chua2005cultural} differences in scene understanding. Individual- and culture-specific AUTO-SUMs could extend the approach to predict such differences in fixation behavior. Such an extension might require collecting scene descriptions from specific individuals or cultural groups, or adapting MLLMs to reproduce the descriptions of a given group.

\section{METHODS}
\label{sec:methods}
\subsection{Stimuli}
We use three image sets with corresponding human fixation data: 36 Winoground images \cite{murlidaran2025eye}, 307 images from the Object and Semantic Images and Eye-tracking (OSIE) dataset \cite{xu2014predicting}, and the 40 OSIE images (OSIE-40) used by Linka et al. \cite{linka2025protracted}.

\subsubsection{Winoground Images}
The Winoground images are taken from Murlidaran and Eckstein (2025) \cite{murlidaran2025eye} (36 images). The images are organized in pairs of real-world scenes, each containing at least one person. The two images of a pair differ only by a small visual change, such as an object's position, an object substitution, or an actor's posture. This change alters the scene's interpretation while largely preserving its low-level saliency. Human fixations were recorded during a scene description task. The data comprise the original 25 subjects per image and an additional 20 subjects per image that we collected, for a total of 45 subjects per image.

\subsubsection{Object and Semantic Images and Eye-tracking (OSIE) Dataset}
The OSIE dataset \cite{xu2014predicting} consists of 700 everyday indoor and outdoor scenes, most of which contain multiple dominant objects rather than a single centered one. All objects are segmented with fine contours and annotated with 12 semantic attributes (e.g., face, text, touched, gazed). We select the subset of 307 images that contain between three and six segmented objects. Human fixations were recorded from 15 subjects during a 3\,s free-viewing task.

\subsubsection{OSIE-40 Developmental Dataset}
Linka et al. \cite{linka2025protracted} used OSIE-40, a 40-image subset of OSIE selected to reliably capture individual differences in semantic salience, with additional pixel masks for persons, heads, and hands. Human fixations were recorded from 6,720 subjects aged 5 to 72 years in a museum eye-tracking booth. Subjects freely viewed each image for 3\,s with the instruction to look at the images in any way they wanted.

\subsection{Eye tracking data collection}

The additional eye-tracking data for the Winoground images were collected using the same apparatus and setup as Murlidaran and Eckstein \cite{murlidaran2025eye}. Eye movements were recorded using an EyeLink 1000+ desktop-mounted eye tracker, sampling at 2,000\,Hz with a spatial resolution of 0.01°. Participants sat 75\,cm from a 19-inch monitor (1,280 × 1,024 pixels), which covered a visual angle of about 26.6° horizontally and 21.8° vertically. The stimulus images had a height of roughly 12.7° of visual angle. To minimize head movements, participants used both a chin and forehead rest. Saccades were identified using a velocity threshold of 22°/s and an acceleration threshold of 4000°/s². Only the left eye was tracked. The experiment was run using SR Research's Experiment Builder software.






\subsection{Fixation prediction models}

\subsubsection{Itti-Koch Saliency Model}
Itti-Koch \cite{itti1998model} is a classical bottom-up saliency model. The image is decomposed into intensity, color-opponent, and Gabor orientation channels across a multi-scale Gaussian pyramid. Center-surround differences within each channel highlight local feature contrast, and the resulting maps are normalized and summed into a single saliency map. We include Itti-Koch as a classical low-level baseline.

\subsubsection{Graph Based Visual Saliency (GBVS)}

GBVS \cite{harel2006graph} is a bottom-up saliency model that relies only on low-level image statistics. The idea of GBVS is to encode an image with low-level features, like luminance, contrast, and orientation. A Markov chain is built over the feature map, with transition probabilities set proportional to the dissimilarity between locations weighted by their spatial proximity, so that its equilibrium distribution accumulates mass at locations that stand out from their surround and produces the saliency map. We include GBVS as a low-level baseline, representing what can be predicted about fixation without any use of high-level semantic representation.

\subsubsection{DeepGaze}

DeepGaze IIE \cite{linardos2021deepgaze} is a supervised deep learning model based on human fixation data to predict where people look. An image is passed through frozen ImageNet-pretrained CNN backbones, and the resulting deep activations are fed to a readout network, combined with a center-bias prior, and finally produce a probabilistic fixation density over the image. The model is optimized for maximum likelihood on fixation data. Unlike other computation models, DeepGaze IIE is trained directly on fixations and serves here as an upper-bound reference for how much fixation variance a state-of-the-art saliency model can capture.

\subsubsection{DeepMeaning}

DeepMeaning \cite{hayes2025deepmeaning} is an image-computable model trained to predict human meaning maps \cite{henderson2017meaning}, where the meaning map is a model that never sees eye-movement data but estimates saliency from local semantic content. In Deepmeaning, each scene is divided into a dense grid of overlapping square patches, and every patch is passed through the frozen vision encoder of a pretrained vision-language transformer (CoCa \cite{yu2022coca}) to obtain an embedding. A linear model, fit once on human meaning ratings and separately for indoor and outdoor scenes, maps each patch embedding to a predicted meaning value. Placing these predictions back at their spatial locations and averaging over the overlap reconstructs the meaning map without any human raters, which we then use directly as a baseline predictor of fixation density. Notably, the model never sees eye-movement data: any correspondence with fixations comes purely from its estimate of local semantic content.

\subsubsection{The AUTO-SUM pipeline}
\label{Methods:AUTOSUM}
The scene understanding map (SUM) developed by Murlidaran and Eckstein \cite{murlidaran2025eye} measures how much an object shapes the understanding of a scene by quantifying how scene descriptions change once that object is removed. The advent of multimodal large language models (MLLMs) that can reason about scene content has made it possible to automate every stage of SUM generation. The automated pipeline (AUTO-SUM) consists of three stages: the Object Removal (OR) stage, the Scene Description (SD) stage, and the Similarity Rating (SR) stage.

\subsubsection*{Automating the Object Removal (OR) Stage}
\begin{enumerate}
\item \textbf{Object list generation:} GPT-4o is prompted to list the names of the objects in the scene, explicitly excluding background elements such as sky, sea, ground, or walls. Removing such elements would cause disproportionate semantic shifts or obvious hallucinations because of their spatial extent. For each scene, we draw two independent responses via multinomial sampling (temperature 1.0) to increase object coverage. GPT-4o then merges the two lists into one, removing duplicated labels.

\item \textbf{Object segmentation and mask preprocessing:} For each object name, the text-prompted segmentation model SAM 3 \cite{carion2025sam} generates segmentation masks with a confidence threshold of 0.4. The masks are then preprocessed in four steps. First, duplicated masks are removed by discarding one mask from every pair whose intersection over union (IoU) exceeds 95\%. Second, following principles of perceptual grouping, spatially proximate masks (distance below 30 pixels) with identical labels are merged into a single mask, so that grouped elements are treated as one entity. Third, masks covering more than 30\% of the image area are discarded to avoid removing large structural components. Finally, each mask is dilated to ensure complete coverage of the object before inpainting.

\item \textbf{Image inpainting:} AUTO-SUM-C uses the closed-source image generation model Nano Banana 2. We overlay the object mask in red on the original image and prompt the model: \textit{Remove the object covered with red masks; ONLY remove the object itself.} AUTO-SUM-O uses MAT \cite{li2022mat}, an open-source transformer-based inpainting model that reconstructs the masked region from its surrounding context. In both cases, each original scene yields one object-removed variant for each retained mask.
\end{enumerate}

\subsubsection*{Automating the Scene Description (SD) Stage}
For each original scene and each of its object-removed variants, we draw 5 descriptions via multinomial sampling from GPT-5.4 (AUTO-SUM-C) or Ovis2 \cite{lu2024ovis} (AUTO-SUM-O).

\subsubsection*{Automating the Similarity Rating (SR) Stage}
All scene descriptions are embedded with the open-source text embedding model Jasper \cite{zhang2024jasper}. Let $D = \{d_1, \dots, d_n\}$ denote the descriptions of the original scene and $D'_o = \{d'_1, \dots, d'_m\}$ the descriptions of the variant in which object $o$ is removed. Sampled descriptions vary from one sample to another, so we compare the two sets as a whole rather than relying on any single description. Specifically, we average the cosine similarity over all $n \times m$ pairs of original and object-removed descriptions. The importance of object $o$ is one minus this average:
\[
\mathrm{SUM}(o) = 1 - \frac{1}{nm} \sum_{j=1}^{n} \sum_{k=1}^{m}
\frac{\mathbf{e}(d_j) \cdot \mathbf{e}(d'_k)}{\lVert \mathbf{e}(d_j) \rVert \, \lVert \mathbf{e}(d'_k) \rVert},
\]
where $\mathbf{e}(\cdot)$ is the Jasper sentence embedding and $n = m = 5$ for AUTO-SUM. If removing an object leaves the descriptions largely unchanged, the average similarity is high and $\mathrm{SUM}(o)$ is close to zero. If removing the object changes the interpretation of the scene, the similarity drops and $\mathrm{SUM}(o)$ increases. The score of each object is assigned to all pixels within its segmentation mask to produce the final AUTO-SUM.

\subsubsection{H-SUM: Human Scene Descriptions with Automated Object Removal and Similarity Rating}
\label{Methods:HSUM}
In the original scene understanding map procedure of Murlidaran and Eckstein \cite{murlidaran2025eye}, all three stages involve human participants. In the Object Removal (OR) stage, objects are digitally removed from each scene one at a time by manual photo editing. In the Scene Description (SD) stage, human participants describe the original scene (the gold standard description) and each object-removed version. In the Similarity Rating (SR) stage, a separate group of human raters judges the similarity between the gold standard and each description of an object-removed scene on a 1--10 scale. In our implementation, only the SD stage is performed by humans. The OR stage reuses the object-removed images generated by the AUTO-SUM-C pipeline, so that H-SUM and AUTO-SUM are computed on identical stimuli. The SR stage replaces human similarity ratings with the embedding-based similarity used in AUTO-SUM. This substitution is supported by Murlidaran and Eckstein \cite{murlidaran2025eye}, who found that similarity scores computed from LLM sentence embeddings agreed with human ratings ($r = 0.70$--$0.72$) about as well as human raters agreed with each other ($r = 0.75$). Consequently, H-SUM and AUTO-SUM differ only in whether the scene descriptions come from humans or from an MLLM.

\subsubsection*{Object Removal (OR) Stage}
We use the object-removed images produced by Nano Banana 2 in the AUTO-SUM-C pipeline. To exclude failed removals, all generated images were verified by 14 human annotators recruited via Prolific. Annotators viewed side-by-side pairs consisting of the original image with the target object highlighted by a red mask and the corresponding object-removed image, and judged whether the target object had been successfully removed. The images were divided into 7 batches of fewer than 300 trials, each evaluated by two independent annotators. Any image flagged as a failure by either annotator was manually reviewed by the researchers, and images with failed removals were discarded. Scenes left with fewer than three valid object-removed variants were removed entirely. The final set comprises 307 original scenes and 1,306 object-removed variants.

\subsubsection*{Scene Description (SD) Stage}
Human descriptions were collected on Prolific from native or highly fluent English speakers with a task approval rate of at least 98\%. Participants received the same instruction used to prompt the MLLMs: \textit{``Make your best guess of what might be happening in this scene in one sentence. Avoid mentioning objects that do not aid in understanding the context of the scene.''} Before the main trials, participants reviewed example descriptions for images not included in the dataset. To prevent memory and carryover effects, we used a between-subjects design. For a scene with $N$ removed objects, the original image and its $N$ variants were distributed across $N+1$ mutually exclusive stimulus sets, and each set was assigned to an independent group of 10 participants. Thus, no participant viewed more than one version of the same scene, and each image received 10 independent descriptions. Participants were not informed that some images had been edited. In total, 227 participants completed 16,665 trials. Participants identified as automated bots or as giving consistently low-effort responses were excluded. We also discarded descriptions whose mean cosine similarity of Jasper embeddings \cite{zhang2024jasper} to the other descriptions of the same image was below 0.4, removing 62 trials (about 0.3\%) and leaving 16,603 valid descriptions.

\subsubsection*{Similarity Rating (SR) Stage}
Instead of human similarity ratings, object importance is computed from the human descriptions with the same mean pairwise cosine similarity of Jasper embeddings as in AUTO-SUM. Here $n$ and $m$ are the numbers of valid human descriptions for the original image and the object-removed variant, which are typically 10 but can vary slightly because of response filtering. The resulting scores are assigned to the corresponding segmentation masks to produce the final H-SUM.

\subsubsection{False discovery rate}

For all analyses involving multiple metric comparisons, p-values were corrected using the Benjamini-Hochberg false discovery rate (FDR) method \cite{benjamini1995controlling}. Statistical significance was evaluated using one-tailed tests.

\section*{ACKNOWLEDGEMENTS}

Research was sponsored by the U.S. Army Research Office and accomplished under cooperative agreement W911NF-19-2-0026 for the Institute for Collaborative Biotechnologies.

\bibliographystyle{unsrt}  
\bibliography{REFERENCES}  


\end{document}